\documentclass[10pt,twocolumn,letterpaper]{article}

\usepackage[pagenumbers]{cvpr} 

\definecolor{cvprblue}{rgb}{0.21,0.49,0.74}
\usepackage[pagebackref,breaklinks,colorlinks,allcolors=cvprblue]{hyperref}
\usepackage{lineno}
\usepackage{fix-cm}
\usepackage{graphicx} 
\usepackage{algorithm}
\usepackage{algorithmic}
\usepackage{multirow}
\usepackage{xspace}
\usepackage{verbatim}
\usepackage{subcaption}
\usepackage{amsmath}
\usepackage{amsthm}
\usepackage{mathrsfs}
\usepackage{amssymb}
\usepackage{dsfont}
\usepackage{pifont}
\usepackage{array}
\usepackage{colortbl}
\usepackage[accsupp]{axessibility}  
\newcommand{\cmark}{\textcolor{red}{\ding{51}}}
\newcommand{\xmark}{\ding{55}}
\newcommand{\baby}{L$^2$\textsc{p-ahil}\xspace}

\def\paperID{1168} 
\def\confName{CVPR}
\def\confYear{2025}

\title{Forming Auxiliary High-confident Instance-level Loss\\to Promote Learning from Label Proportions}

\author{
Tianhao Ma\textsuperscript{1,2,}\thanks{Equal contribution.} , Han Chen\textsuperscript{1,2,}\footnotemark[1] , Juncheng Hu\textsuperscript{1,2}, Yungang Zhu\textsuperscript{1,2}, Ximing Li\textsuperscript{1,2,}\thanks{Corresponding author.} \vspace{2pt}\\
\textsuperscript{1}College of Computer Science and Technology, Jilin University, China\\
\textsuperscript{2}Key Laboratory of Symbolic Computation and Knowledge Engineering of Ministry of Education,\\ Jilin University, China\\
{
    \tt\small \{matianhao2120, 2002chenhan\}@gmail.com, 
    \{jchu, zhuyungang\}@jlu.edu.cn,
    liximing86@gmail.com
}
}

\begin{document}
\maketitle
\begin{abstract}
    Learning from label proportions (LLP), \ie, a challenging weakly-supervised learning task, aims to train a classifier by using bags of instances and the proportions of classes within bags, rather than annotated labels for each instance. Beyond the traditional bag-level loss, the mainstream methodology of LLP is to incorporate an auxiliary instance-level loss with pseudo-labels formed by predictions. Unfortunately, we empirically observed that the pseudo-labels are are often inaccurate due to over-smoothing, especially for the scenarios with large bag sizes, hurting the classifier induction. To alleviate this problem, we suggest a novel LLP method, namely Learning from Label Proportions with Auxiliary High-confident Instance-level Loss (\baby). Specifically, we propose a dual entropy-based weight (DEW) method to adaptively measure the confidences of pseudo-labels. It simultaneously emphasizes accurate predictions at the bag level and avoids overly smoothed predictions. We then form high-confident instance-level loss with DEW, and jointly optimize it with the bag-level loss in a self-training manner. The experimental results on benchmark datasets show that \baby can surpass the existing baseline methods, and the performance gain can be more significant as the bag size increases. The implementation of our method is available at \url{https://github.com/TianhaoMa5/LLP-AHIL}.
\end{abstract}
\section{Introduction}
\label{introduction}

During the past decades, supervised learning has been proven to gain significant success in diverse real-world applications, in which one can easily collect loads of training instances with precise supervision~\citep{lin2023rich,lin2024consistent123}. Unfortunately, in many scenarios, it is challenging to collect a sufficient number of precisely labeled training instances due to various difficulties such as high annotation costs~\citep{la2023learning,ramos2023lightweight} and privacy protection concerns~\citep{hernandez2018fitting}. Consequently, training instances with incomplete, inexact, and inaccurate supervision are often available only~\citep{zhou2018brief}. The demand for learning with such weak supervision facilitates the development of \textbf{W}eakly-\textbf{S}upervised \textbf{L}earning (\textbf{WSL})~\citep{book:Sugiyama+etal:2022}, ranging many specific paradigms, \eg, semi-supervised learning~\citep{li2021semi,zheng2022simmatch}, partial label learning~\citep{lv2020progressive,li2023learning}, and positive-unlabeled learning~\citep{guo2020positive,hsieh2019classification}, to name just a few.

\begin{figure}[t!]
    \centering
    \includegraphics[scale=0.38]{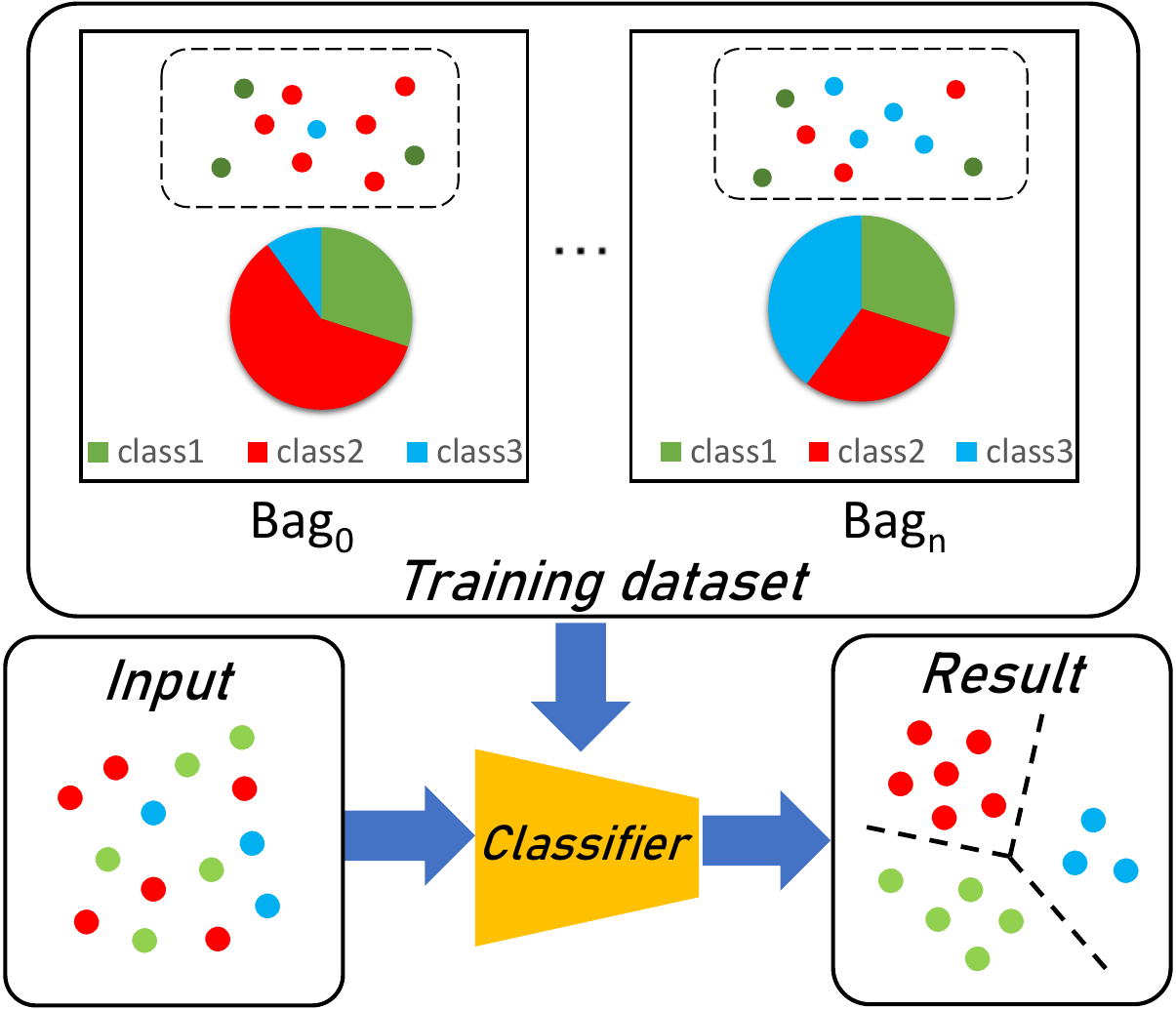}
    \caption{Before training, the data is grouped and only the proportion information is retained. Classifier training is performed using unlabeled data and proportions in bag units, and finally the instance classification task is completed.}    
    \label{intro_llp}
\end{figure}

\begin{figure*}[t!]
    \centering
    \subcaptionbox{}{\includegraphics[scale=0.43]{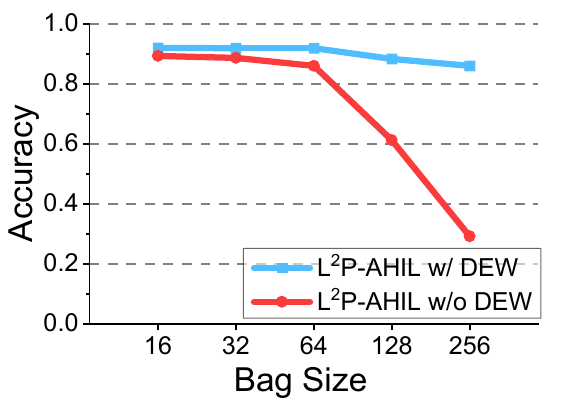}\label{fig:cifar10-acc}}
    \subcaptionbox{}{\includegraphics[scale=0.43]{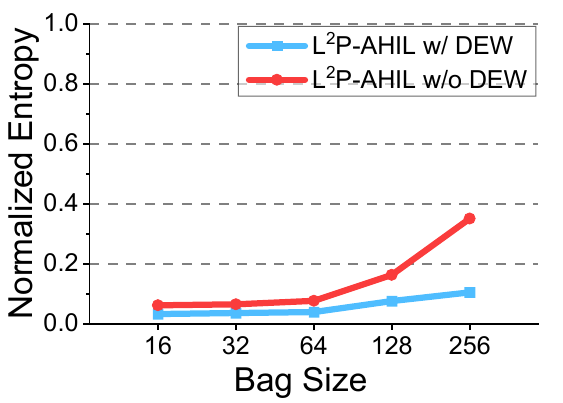}\label{fig:cifar10-ent}}
    \subcaptionbox{}{\includegraphics[scale=0.43]{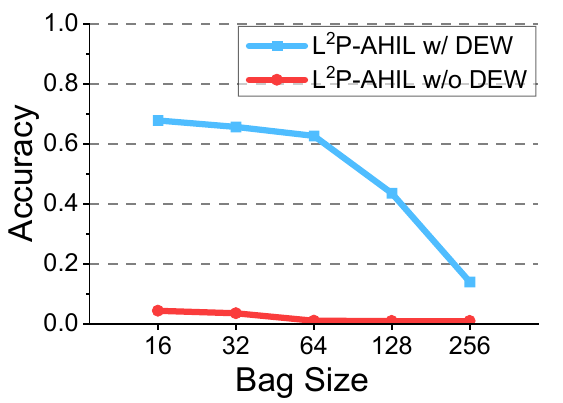}\label{fig:cifar100-acc}}
    \subcaptionbox{}{\includegraphics[scale=0.43]{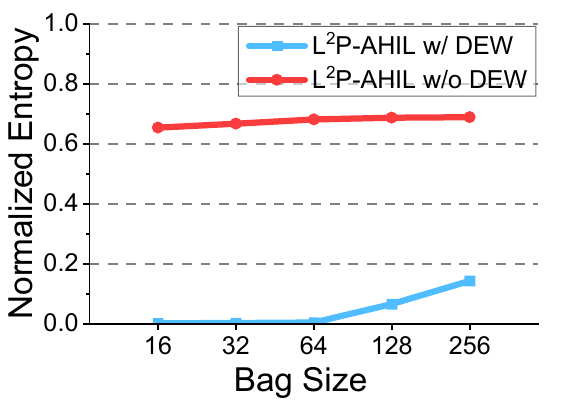}\label{fig:cifar100-ent}}
    \caption{Results of preliminary experiments with different bag sizes on CIFAR-10 and CIFAR-100 (500 epochs). (a)(c) Averaged accuracy of pseudo-labels on CIFAR-10 and CIFAR-100. (b)(d) Averaged normalized entropy of pseudo-labels on CIFAR-10 and CIFAR-100. Higher normalized entropy values imply more smoothing results. Table~\ref{tab:ablation} shows the adverse effects caused by these inaccurate pseudo-labels.}    
    \label{fig:motivation}
\end{figure*}

Recently, an emerging paradigm of WSL, known as \textbf{L}earning from \textbf{L}abel \textbf{P}roportions (\textbf{LLP}) as shown in Fig.~\ref{intro_llp}, has gained increasing attention within the machine learning community~\citep{zhang2022learning,asanomi2023mixbag,Havaldar2023Learning,brahmbhatt2023pac}, and it has been broadly applied across various computer vision domains such as remote sensing image categorization~\citep{li2015alter,ding2017learning} and medical image processing~\citep{bortsova2018deep}. Technically, LLP, as its name suggests, refers to the classification problem where a certain number of training instances are grouped into many bags with the corresponding proportion of classes, rather than annotated labels for each instance. Naturally, such sparse and inexact supervision from LLP causes significant difficulties in inducing effective classifiers for predicting labels of instances. 

To handle the task of LLP, the basic method is to induce the classifier by optimizing the bag-level loss, measuring the difference between the ground-truth class proportion in a bag and the predicted class proportion formed by predictions of instances from the bag~\citep{yu2014on}. 
As suggested by~\citep{Liu2019Learning,liu2021two}, optimizing the bag-level loss only can not guarantee inducing effective classifiers. Therefore, the mainstream methodology of LLP becomes to incorporate an auxiliary instance-level loss with pseudo-labels, and jointly optimize it with the bag-level loss in a self-training manner~\citep{liu2021two,liu2022self-llp,Matsuo2023Learning,dulac-arnold2019deep}. However, in our preliminary experiments, we empirically observed that the pseudo-labels generated by this kind of methods are often inaccurate and even meaningless, especially for the scenarios with large bag sizes. As illustrated in Fig.~\ref{fig:motivation}, our early empirical results indicated that the pseudo-labels are inaccurate and tend to be overly smoothed, quantitatively measured by entropy values, as the bag size increases. Such low-quality pseudo-labels hurt the classifier induction, resulting in performance degradation. 

To solve this problem, in this paper, we propose a novel LLP method, namely \textbf{L}earning from \textbf{L}abel \textbf{P}roportions with \textbf{A}uxiliary \textbf{H}igh-confident \textbf{I}nstance-level \textbf{L}oss (\textbf{\baby}). Specifically, we suggest a dual entropy-based weight~(DEW) method to adaptively evaluate the confidence of pseudo-labels from both bag and instance levels. At the bag level, it emphasizes accurate class proportion predictions measured the difference between entropy values of ground-truth proportions and predicted proportions from the class-pivoted perspective. At the instance level, it avoids overly smoothed predictions of instances, directly measured by the entropy values. We then form a high-confident instance-level loss with DEW, and jointly optimize it with the bag-level loss in a self-training manner. The experimental results on benchmark datasets demonstrate that \baby can outperform the existing baseline methods, and the performance gain can be more significant as the bag size increases.

In a nutshell, the major contributions of this paper are outlined as follows:

\begin{itemize}
    \item We quantitatively evaluated the problem of inaccurate and meaningless pseudo-labels of the existing LLP art, as the big size increases.  
    
    \item To solve this problem, we propose a novel LLP method named \baby, which adaptively measures the confidences of pseudo-labels.
    
    \item We conduct extensive experiments to empirically validate the effectiveness of \baby even with larger bag sizes.
\end{itemize}

\section{Related Work}
\label{related work}

\paragraph{Learning from label proportions.}
~\citep{yu2014on} theoretically demonstrated that an instance-level classifier can be trained using bag-level proportion information. Building on this theory, DLLP~\citep{ardehaly2017co} proposed a bag-level loss that aligns the average of all instances' predictions within a bag with the target distribution. Many methods have been developed based on this loss. LLP-GAN~\citep{Liu2019Learning} introduced a generator to produce fake instances, increasing the class count to \( K+1 \), enabling the model to classify these fake instances as the $ (K+1)^\text{th} $ class. LLP-VAT~\citep{tsai2020learning} uses a consistency regularization loss to maintain uniform predictions for an instance across various data augmentations.~\citep{nandy2022domain,yang2021two} introduced self-training with a contrastive learning loss to aid training. These methods incorporate self-supervised representation learning, but without instance-level supervised training, it is difficult to train a well-performing instance-level classifier. 

To improve instance-level classification performance, some approaches employ the model's predictions as pseudo-labels for self-training. SELF-LLP~\citep{liu2022self-llp} employs the model's predictions from different epochs as pseudo-labels for training. ROT and PLOT~\citep{dulac-arnold2019deep,liu2021two} utilize pseudo-labels generated through optimal transport for training. In the study by~\citep{Matsuo2023Learning}, they employed an online pseudo-labeling method with regret minimization. However, due to the significant smoothing of model predictions, especially with larger bag sizes, it naturally leads to the generation of numerous incorrect pseudo-labels. In our approach, we utilize pseudo-labels for self-training and adaptively adjust the weight of each instance's pseudo-label based on both bag-level and instance-level entropy values. This prevents unreliable pseudo-labels from compromising the model's performance.

From another perspective, the works of~\citep{Saket2021Learnability,Saket2022Algorithms, qua2008est, patrini2014no, zhang2022learning, kobayashi2022learning, busa-fekete2023easy} are grounded in statistical learning theory. However, these approaches have significant limitations. For instance, the method proposed in~\citep{Saket2021Learnability,Saket2022Algorithms} heavily relies on the assumption that an example can belong to multiple bags. On the other hand,~\citep{busa-fekete2023easy} introduced an unbiased estimator, but their method suffered from severe overfitting due to negative empirical loss.

\begin{figure*}[t!]
    \centering
    \begin{subfigure}[b]{0.22\textwidth}
        \centering
        \includegraphics[scale=0.39]{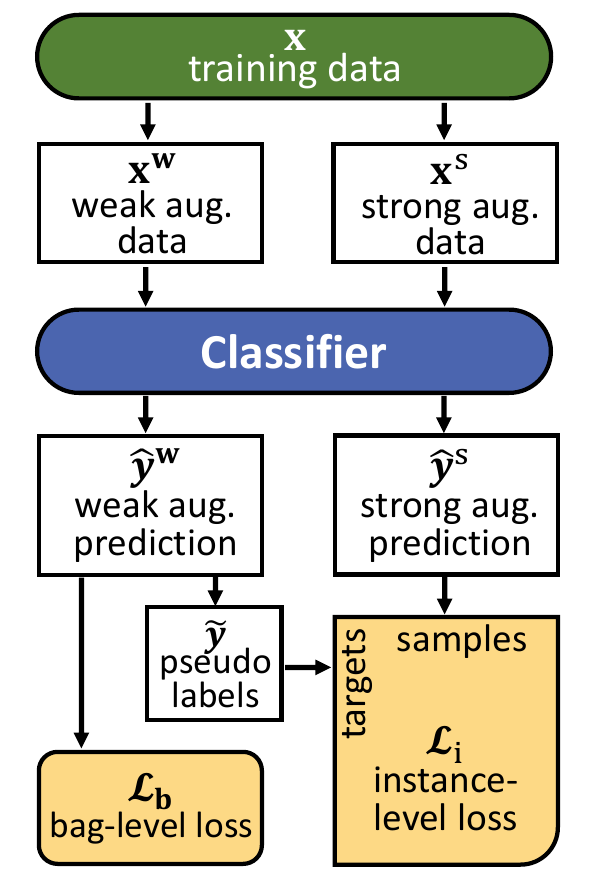}
        \caption{Model pipeline}
        \label{fig:pipeline}
    \end{subfigure}
    \hfill
    \begin{subfigure}[b]{0.77\textwidth}
        \centering
        \includegraphics[scale=1.10]{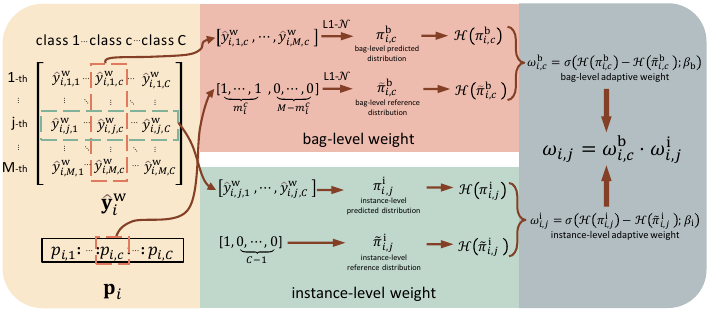}
        \caption{The process of DEW calculation}
        \label{fig:DEW-process}
    \end{subfigure}
    \caption{
        (a) The plot of the model's pipeline with the ``Weak and Strong Augmentation" strategy (aug. denotes augmentation). When calculating the instance-level loss \( \mathcal{L}_\mathrm{i} \), \( \mathbf{\hat y}^\mathrm{s} \) is used as input samples and \( \mathbf{\tilde y} \) as targets. Adaptive weight components are omitted for visual clarity. 
        (b) The plot of the detail process of DEW. The process involves: 1) acquiring predicted and reference distributions at both bag and instance levels (red and green pathways, respectively), 2) calculating entropy for these distributions, and 3) determining adaptive weights through a mapping function that integrates information from both levels. $\text{L1-}\mathcal{N}$ indicates L1-Normalization operations.
    }
    \label{fig:method}
\end{figure*}

\paragraph{Adaptive weighting techniques in weakly-supervised learning.}
\label{awt}
In the domain of semi-supervised learning, many methods utilize an adaptive weighting approach for pseudo-labels.~\citep{lai2022smoothed} adaptively adjust the weight of each class by estimating the learning difficulty of each class.
Influence~\citep{Ren2020Not} introduces the concept of adjusting weights using an algorithm based on the influence function. SoftMatch~\citep{Chen2023SoftMatch} employs a truncated Gaussian function to weight samples according to their confidence levels.~\citep{iscenyac2019label} adaptively adjust the weight of each instance based on the entropy of the pseudo-labels for each instance. However, many adaptive weighting strategies are not specifically tailored for LLP. Directly applying these methods to LLP may not yield optimal performance, as demonstrated in Sec. \ref{experiments}. In our method, we measure the confidences of pseudo-labels based on the DEW method, adaptively adjusting the weight of each instance accordingly. 
\section{Proposed \baby Method}
\label{method}

In this section, we introduce the proposed LLP method named \baby.

\paragraph{Task formulation of LLP.}
We now introduce essential notations to formulate the LLP task.  
Typically, we are given an LLP training dataset of \(N\) proportion-labeled bags \(\mathcal{D} = \{(\mathbf{B}_i, \mathbf{p}_i)\}_{i=1}^{N}\), where each \(\mathbf{B}_i = \{\mathbf{x}_{i,j}\}_{j=1}^{M}\) denotes a bag of \(M\) instances represented by \(\mathbf{x}\). Its corresponding proportion label \(\mathbf{p}_i \in \triangle^{C-1}\) denotes the proportion of \(C\) pre-defined classes in the bag. The value of \(m_i^c = p_{i,c}M\) represents the number of instances of class \(c\) in $\mathbf{B}_i$. Generally, all bags are disjoint, \ie, $\mathbf{B}_i \cap \mathbf{B}_k = \varnothing, \forall i \neq k$. The goal of LLP is to induce a classifier over $\mathcal{D}$ with weakly supervised signals, \ie, the bag-level label proportion only.

\subsection{Overview}

Overall speaking, our \baby method is denoted as $f_\mathbf{W}(\cdot)$, parameterized by $\mathbf{W}$, which ingests an instance $\mathbf{x}$ and outputs its prediction $\mathbf{\hat y}=f_{\mathbf{W}}(\mathbf{x})$.  We follow the methodology of incorporating an auxiliary instance-level loss with pseudo-labels. For the instance-level loss, we adopt the commonly used ``Weak and Strong Augmentation'' strategy in the semi-supervised learning domain~\citep{Sohn2020FixMatch} to enhance the model's representational learning capabilities. For each instance \(\mathbf{x}\), we transform it into a weakly augmented version \(\mathbf{x}^\mathrm{w}\) and a strongly augmented version \(\mathbf{x}^\mathrm{s}\), where \(\mathbf{x}^\mathrm{w}\) is specifically used to form the corresponding pseudo-label and \(\mathbf{x}^\mathrm{s}\) is used as a training sample. The predictions for them are represented by \(\mathbf{\hat y}^\mathrm{w} = f_{\mathbf{W}}(\mathbf{x}^\mathrm{w})\) and \(\mathbf{\hat y}^\mathrm{s} = f_{\mathbf{W}}(\mathbf{x}^\mathrm{s})\), respectively.
With the predictions of all instances, we can formulate the objective of \baby with respect to $\mathbf{W}$ as follows:
\begin{equation}
\label{object}
\mathcal{L}(\mathbf{W}) = \mathcal{L}_{\rm{b}}(\mathbf{p},\mathbf{\hat y}^{\mathrm{w}}) + \lambda \mathcal{L}_{\rm{i}}(\mathbf{\tilde y},\mathbf{\hat y}^\mathrm{s}),
\end{equation}
where $\mathcal{L}_{\rm{b}}(\mathbf{p},\mathbf{\hat y}^{\mathrm{w}})$ is the bag-level proportion loss; $\mathcal{L}_{\rm{i}}(\mathbf{\tilde y},\mathbf{\hat y}^\mathrm{s})$ is the instance-level self-training loss and $\mathbf{\tilde y}$ denotes the pseudo-label; $\lambda$ is coefficient parameter used to balance the two losses. The overall pipeline of the proposed model is illustrated in Fig.~\ref{fig:pipeline}.

\paragraph{Bag-level proportion loss.}
Following~\citep{ardehaly2017co}, \baby continues to utilize this loss for bag-level proportion learning. It is defined as the cross-entropy between prior and posterior proportions which can be formulated as follows:
\begin{equation}
\label{loss_b}
    \mathcal{L}_{\rm b}(\mathbf p, \mathbf{\hat y}^\mathrm{w}) = \frac1N \sum_{i=1}^N\mathcal{H}(\mathbf{p}_i, \mathbf{\bar p}_i),
\end{equation}
where $\mathbf{\bar p}_i = \left[\bar p_{i,c}\right]_{c=1}^C$ denotes the predicted  proportion in $\mathbf{B}_i$, and $\mathcal{H}(\cdot, \cdot)$ denotes the cross-entropy loss. The value of $\bar p_{i,c}$ is computed by averaging the predicted probabilities of class $c$ across all instances in $\mathbf{B}_i$. It can be represented by the formula:
\begin{equation}
\label{bar_p}
\bar p_{i,c} = \frac1M \sum_{j=1}^{M} \hat y^{\mathrm{w}}_{i,j,c},
\end{equation}
where $\hat y^{\mathrm{w}}_{i,j,c}$  represents the predicted probability of class $c$ for $\mathbf{x}_{i,j}^{\mathrm{w}}$. 

\paragraph{Instance-level self-training loss.}
We define the pseudo-labels $\mathbf{\tilde y}  $ as the hardened version of \(\mathbf{\hat y}^\mathrm{w}\). This means that for each \( \mathbf{x}_{i,j} \), its pseudo label $\mathbf{\tilde y}_{i,j} = [\tilde{y}_{i,j,c}]_{c=1}^C$ has one class with a probability of $1$, while the probabilities for all other classes are set to $0$. Formally, this can be expressed as:
\begin{equation}
\label{pseudo-labels}
\tilde y_{i,j,c} = 
    \begin{cases}
	1, &\text{if }c = \arg \max(\mathbf{\hat y}^{\mathrm{w}}_{i,j}), \\
	0, &\text{otherwise}.
    \end{cases}
\end{equation}
Using the strongly augmented predictions $\mathbf{\hat y}^\mathrm{s}$ as training samples and the pseudo labels $\mathbf{\tilde y}  $ as training targets, we employ cross entropy as the loss function to calculate the instance-level loss:
\begin{equation}
\label{loss_i}
\mathcal{L}_{\rm{i}}(\mathbf{\tilde y},\mathbf{\hat y}^{\mathrm{s}}) = \frac1{N\cdot M} \sum_{i=1}^N \sum_{j=1}^{M} \omega_{i,j}\mathcal{H}(\mathbf{\tilde y}_{i,j}, \mathbf{\hat y}_{i,j}^{\rm s}),
\end{equation}
where $\omega_{i,j}$ denotes the adaptive weight for $\mathbf{x}_{i,j}$ according to its confidence. How to measure this confidence is the key to \baby, and we will describe the details in the following parts.

\subsection{Dual Entropy-based Weight}
\label{subsec:DEW}

Based on the assumption that lower entropy values imply high confidence in LLP, we propose a DEW weight for computing $\omega_{i,j}$. Specifically, we consider the confidence from both bag and instance levels. Accordingly, we design $\omega_{i,j}$ as a product of a bag-level adaptive weight \(\omega^{\rm b}\) and an instance-level adaptive weight \(\omega^{\rm i}\) as follows:
\begin{equation}
\label{omega}
    \omega_{i,j} = \omega^{\rm{b}}_{i,c} \cdot\omega^{\rm{i}}_{i,j},
\end{equation}
where $\omega_{i,j}$ represents the adaptive weight for $\mathbf{x}_{i,j}$ and here, $c=\arg\max\mathbf{\hat y}^{\rm{w}}_{i,j}$ denotes the class with the highest probability. We calculate \(\omega^{\rm{b}}_{i,c}\) and \(\omega^{\rm{i}}_{i,j}\) based on the entropy differences between the predicted distribution and the reference distribution at the bag and instance level, respectively. The detailed process of DEW is shown in Fig.~\ref{fig:DEW-process}.

\subsubsection{Bag-level Adaptive Weight}

At the bag level, the predicted distribution for class $c$ is constructed by aggregating the predictions from all instances within bag $\mathbf{B}_i$. Specifically, we first collect the unnormalized predictions $\left[\hat{y}^{\mathrm{w}}_{i,j,c}\right]_{j=1}^M$ into a vector denoted as ${\mathbf{\pi}^\prime}_{i,c}^{\mathrm{b}}$. To obtain a valid probability distribution (i.e., summing to $1$) at the bag level, we apply L1 normalization to ${\mathbf{\pi}^\prime}_{i,c}^{\mathrm{b}}$, resulting in the bag-level predicted distribution $\mathbf{\pi}_{i,c}^{\mathrm{b}} = {\mathbf{\pi}^\prime}_{i,c}^{\mathrm{b}} / \lVert {\mathbf{\pi}^\prime}_{i,c}^{\mathrm{b}} \rVert_1$. 

Based on practical knowledge, we define bag-level reference distribution as \(\mathbf{\tilde{\pi}}^\mathrm{b}_{i,c}\), which is composed as follows: there are \(m_i^c\) instances predicted as class \(c\) with the highest probability, i.e., \(1/m_i^c\), while the remaining \((M-m_i^c)\) instances have the lowest probability, i.e., 0. It can be expressed as follows:
\begin{equation}
  \begin{aligned}
    \mathbf{\tilde \pi}^\mathrm{b}_{i,c}  &= 
    \text{L1-Normalization}\big(\big[\underbrace{\vphantom{1} 1,\cdots,1}_{m_i^c}  ,\underbrace{\vphantom{1} 0,\cdots, 0}_{M-m_i^c}\big]\big) \\
    &= \bigg[ \underbrace{\vphantom{\frac1{n_{i,c}}} \frac1{m_i^c},\cdots,\frac1{m_i^c}}_{m_i^c}  ,\underbrace{\vphantom{\frac1{m_i^c}} 0,\cdots, 0}_{M-m_i^c} \bigg].
  \end{aligned}
\end{equation}
The entropy of the bag-level reference distribution can then be calculated as $\mathcal{H}(\mathbf{\tilde{\pi}}^\mathrm{b}_{i,c}) = \log m_i^c$. As the predicted distribution approaches the reference distribution, the adaptive weight tends toward 1; conversely, it tends toward 0 when the distributions diverge. Therefore, we introduce $\sigma(x;\beta)=\exp\left(-\frac{x^2}{\beta}\right)$ as a mapping function, which allows us to derive the bag-level adaptive weight:
\begin{equation}
\begin{aligned}
\label{weight_c}
\omega^{\mathrm{b}}_{i,c} 
&=\sigma\left (\mathcal{H}(\mathbf{\pi}^\mathrm{b}_{i,c}) - \mathcal{H}(\mathbf{\tilde{\pi}}^\mathrm{b}_{i,c}); \beta_{\mathrm{b}}\right ) \\
&= \exp\left\{ -\frac{\left[ \mathcal{H}(\mathbf{\pi}^\mathrm{b}_{i,c}) - \mathcal{H}(\mathbf{\tilde{\pi}}^\mathrm{b}_{i,c}) \right]^2}{\beta_{\mathrm{b}}}\right\} \\
&= \exp\left\{ -\frac{\left[ \mathcal{H}(\mathbf{\pi}^\mathrm{b}_{i,c}) - \log m_i^c \right]^2}{\beta_{\mathrm{b}}}\right\},
\end{aligned}
\end{equation}
where \(\beta_{\rm b} > 0\) is a hyper parameter that controls the degree of smoothness of the mapping function.

\subsubsection{Instance-level Adaptive Weight}

At the instance level, the predicted distribution is derived from the prediction results of individual instances. Specifically, for each instance \(\mathbf{x}_{i,j}\), we define instance-level predicted distribution \(\mathbf{\pi}_{i,j}^{\mathrm{i}} = \left[\hat{y}^{\mathrm{w}}_{i,j,c}\right]_{c=1}^C\) to measure the confidence of the instance.

For the instance-level reference distribution, we consider a label to have high confidence when its predicted probabilities are concentrated on a single class. Therefore, we define the instance-level reference distribution as \(\mathbf{\tilde{\pi}}_{i,j}^{\mathrm{i}}\), where only one class has a probability of 1, and the probabilities of all other classes are set to 0, \ie,
\begin{equation}
  \begin{aligned}
    \mathbf{\tilde{\pi}}^{\mathrm{i}}_{i,j} &=
    \big[1, \underbrace{\vphantom{0} 0, \ldots, 0}_{C-1}\big].
  \end{aligned}
\end{equation}
It is straightforward to calculate that the entropy of instance-level reference distribution is $0$, \ie, $\mathcal{H}(\mathbf{\tilde \pi}^\mathrm{i}_{i,j})=0$. With the same mapping function as the bag level, the instance-level adaptive weight is obtained with a hyper parameter $\beta_{\mathrm{i}} > 0$:
\begin{equation}
\begin{aligned}
\label{weight_i}
    \omega^\mathrm{i}_{i,j}
    &=\sigma\left (\mathcal{H}(\mathbf{\pi}^\mathrm{i}_{i,j}) - \mathcal{H}(\mathbf{\tilde \pi}^\mathrm{i}_{i,j}); \beta_{\mathrm{i}}\right ) \\
    &=\exp\left\{ -\frac{\left[ \mathcal{H}(\mathbf{\pi}^\mathrm{i}_{i,j}) - \mathcal{H}(\mathbf{\tilde \pi}^\mathrm{i}_{i,j}) \right]^2}{\beta_{\mathrm{i}}}\right\} \\
    &= \exp \left[ -\frac {\mathcal{H}(\mathbf{\pi}_{i,j}^\mathrm{i})^2} {\beta_{\mathrm{i}}}  \right].
\end{aligned}
\end{equation}

\section{Experiments}
\label{experiments}

\subsection{Experimental Setup}

\paragraph{Datasets.}
\label{Sec:datasets}
We assessed \baby on five benchmark datasets for LLP: Fashion-MNIST~\citep{xiao2017fashion}, SVHN~\citep{netzer2011reading}, CIFAR-10, CIFAR-100~\citep{krizhevsky2009learning} and Mini-ImagNet~\citep{vinyals2016matching}. Fashion-MNIST is a dataset consisting of 60,000 training and 10,000 test examples. Each example is a 28x28 gray-scale image, representing 10 categories of fashion items. The SVHN dataset is composed of 32x32 RGB images of digits, featuring 73,257 training examples and 26,032 testing examples, along with an additional 531,131 training samples that we do not utilize in our experiments. CIFAR-10 and CIFAR-100 datasets each contain 50,000 training and 10,000 testing examples, with every example being a 32x32 color natural image. They are categorized into 10 and 100 classes, respectively. MiniImageNet is a subset of the ImageNet dataset, consisting of 100 selected classes. Each class contains 80 images for training and 20 images for testing, with all images resized to 64x64. 

\paragraph{Bag generation.}
For each dataset, with a specified bag size \(M\), we randomly select \(M\) samples from the training set $\mathcal{D}$ to form a bag \(\mathbf{B}_i\). The samples in distinct bags do not overlap. We compute the class proportion information within each bag to guide the training, but real labels are not used during training. Following prior studies, we choose \(M\) from the set \{16, 32, 64, 128, 256\}. As every dataset contains an equal number of samples across classes, the bags created through this method exhibit relatively balanced proportions, leading to more challenging LLP tasks, as evidenced by~\citep{yu2014on}.
\begin{table*}[ht]
    \caption{Classification accuracy (mean±std) on CIFAR-10, CIFAR-100, SVHN, Fashion-MNIST and MiniImageNet for different bag sizes. The highest accuracy is highlighted in bold. The symbol ``*'' indicates results for FLMm are reproduced directly from its original publication~\citep{yang2021two}, while the dash symbol ``-'' signifies either the inapplicability of LLPFC to CIFAR-100 or the absence of reported results for SVHN, Fashion-MNIST, and MiniImageNet in FLMm's experiments.}
    \label{tab:result}
    \centering
    \fontsize{9.4pt}{9.4pt}\selectfont
    \begin{tabular}{clcccccc}
    \toprule
    \multicolumn{1}{c}{\multirow{2}[1]{*}{Dataset}} & \multicolumn{1}{c}{\multirow{2}[1]{*}{Model}} & \multicolumn{5}{c}{Bag Size} & \multirow{2}[1]{*}{{\parbox{1.5cm}{\centering Fully\\Supervised}}}  \\
\cmidrule{3-7}      &  & \multicolumn{1}{c}{16} & \multicolumn{1}{c}{32} & \multicolumn{1}{c}{64} & \multicolumn{1}{c}{128} & \multicolumn{1}{c}{256} \\
    \midrule
    \multicolumn{1}{c}{\multirow{7}[1]{*}{CIFAR-10}} & LLPFC & 84.10\tiny±0.19 & \multicolumn{1}{l}{71.70\tiny±0.78} & \multicolumn{1}{l}{52.71\tiny±0.36} & \multicolumn{1}{l}{20.78\tiny±0.70} & \multicolumn{1}{l}{18.79\tiny±0.21} & \multicolumn{1}{c}{\multirow{7}[1]{*}{96.05\tiny±0.33}} \\
      & DLLP & 91.59\tiny±1.52 & \multicolumn{1}{l}{88.61\tiny±0.90} & \multicolumn{1}{l}{79.76\tiny±1.45} & \multicolumn{1}{l}{64.95\tiny±0.01} & \multicolumn{1}{l}{44.87\tiny±1.13} \\
      & LLP-VAT & 91.80\tiny±0.08 & \multicolumn{1}{l}{89.11\tiny±0.22} & \multicolumn{1}{l}{78.75\tiny±0.46} & \multicolumn{1}{l}{63.89\tiny±0.19} & \multicolumn{1}{l}{46.93\tiny±0.71} \\
      & ROT & 94.86\tiny±0.68 & \multicolumn{1}{l}{94.34\tiny±0.65} & \multicolumn{1}{l}{93.97\tiny±0.96} & \multicolumn{1}{l}{92.23\tiny±0.81} & \multicolumn{1}{l}{63.10\tiny±0.84} \\
      & SoftMatch & \textbf{95.24\tiny±0.12} & \multicolumn{1}{l}{\textbf{95.25\tiny±0.14}} & \multicolumn{1}{l}{94.23\tiny±0.18} & \multicolumn{1}{l}{\textbf{93.87\tiny±0.22}} & \multicolumn{1}{l}{48.20\tiny±0.26} \\
      & FLMm* & \multicolumn{1}{l}{92.34} & \multicolumn{1}{l}{92.00} & \multicolumn{1}{l}{91.74} & \multicolumn{1}{l}{91.54} & \multicolumn{1}{l}{91.29} \\
      & \baby & 94.96\tiny±0.13 & \multicolumn{1}{l}{95.00\tiny±0.11} & \multicolumn{1}{l}{\textbf{94.58\tiny±0.21}} & \multicolumn{1}{l}{93.64\tiny±0.20} & \multicolumn{1}{l}{\textbf{92.88\tiny±0.53}} \\
    \midrule
    \multicolumn{1}{c}{\multirow{7}[1]{*}{CIFAR-100}} & LLPFC & - & \multicolumn{1}{c}{-} & \multicolumn{1}{c}{-} & \multicolumn{1}{c}{-} & \multicolumn{1}{c}{-} & \multicolumn{1}{c}{\multirow{7}[1]{*}{79.89\tiny±0.14}} \\
      & DLLP & 71.28\tiny±1.56 & \multicolumn{1}{l}{69.92\tiny±2.86} & \multicolumn{1}{l}{53.58\tiny±1.60} & \multicolumn{1}{l}{25.86\tiny±2.15} & \multicolumn{1}{l}{8.82\tiny±0.94} \\
      & LLP-VAT & 73.85\tiny±0.22 & \multicolumn{1}{l}{71.62\tiny±0.07} & \multicolumn{1}{l}{65.31\tiny±0.33} & \multicolumn{1}{l}{37.36\tiny±0.63} & \multicolumn{1}{l}{2.79\tiny±0.67} \\
      & ROT & 72.74\tiny±0.08 & \multicolumn{1}{l}{69.31\tiny±0.22} & \multicolumn{1}{l}{17.48\tiny±0.86} & \multicolumn{1}{l}{11.02\tiny±0.79} & \multicolumn{1}{l}{2.86\tiny±1.11} \\
      & SoftMatch & \textbf{80.14\tiny±0.12} & \multicolumn{1}{l}{2.40\tiny±0.15} & \multicolumn{1}{l}{2.04\tiny±0.10} & \multicolumn{1}{l}{2.12\tiny±0.13} & \multicolumn{1}{l}{1.98\tiny±0.20} \\
      & FLMm* & \multicolumn{1}{l}{66.16} & \multicolumn{1}{l}{65.59} & \multicolumn{1}{l}{64.07} & \multicolumn{1}{l}{61.25} & \multicolumn{1}{l}{\textbf{57.10}} \\
      & \baby & 78.65\tiny±0.28 & \multicolumn{1}{l}{\textbf{77.30\tiny±0.50}} & \multicolumn{1}{l}{\textbf{76.52\tiny±0.23}} & \multicolumn{1}{l}{\textbf{72.21\tiny±0.37}} & \multicolumn{1}{l}{23.56\tiny±2.13} \\
    \midrule
    \multicolumn{1}{c}{\multirow{7}[1]{*}{SVHN}} & LLPFC & 93.04\tiny±0.21 & \multicolumn{1}{l}{23.26\tiny±0.63} & \multicolumn{1}{l}{21.28\tiny±0.23} & \multicolumn{1}{l}{20.54\tiny±0.37} & \multicolumn{1}{l}{19.58\tiny±0.09} & \multicolumn{1}{c}{\multirow{7}[1]{*}{97.77\tiny±0.03}}\\
      & DLLP & 96.90\tiny±0.50 & \multicolumn{1}{l}{96.93\tiny±0.23} & \multicolumn{1}{l}{96.64\tiny±0.32} & \multicolumn{1}{l}{95.51\tiny±0.04} & \multicolumn{1}{l}{94.34\tiny±0.12} \\
      & LLP-VAT & 96.88\tiny±0.03 & \multicolumn{1}{l}{96.68\tiny±0.01} & \multicolumn{1}{l}{96.38\tiny±0.10} & \multicolumn{1}{l}{95.29\tiny±0.17} & \multicolumn{1}{l}{92.18\tiny±0.29} \\
      & ROT & 95.54\tiny±0.10 & \multicolumn{1}{l}{94.78\tiny±0.13} & \multicolumn{1}{l}{96.75\tiny±0.11} & \multicolumn{1}{l}{26.00\tiny±0.43} & \multicolumn{1}{l}{12.15\tiny±0.57} \\
      & SoftMatch & 22.39\tiny±0.11 & \multicolumn{1}{l}{19.68\tiny±0.13} & \multicolumn{1}{l}{19.60\tiny±0.12} & \multicolumn{1}{l}{19.64\tiny±0.14} & \multicolumn{1}{l}{19.57\tiny±0.16} \\
      & FLMm* & - & \multicolumn{1}{c}{-} & \multicolumn{1}{c}{-} & \multicolumn{1}{c}{-} & \multicolumn{1}{c}{-} \\
      & \baby & \textbf{97.91\tiny±0.02} & \multicolumn{1}{l}{\textbf{97.88\tiny±0.01}} & \multicolumn{1}{l}{\textbf{97.74\tiny±0.06}} & \multicolumn{1}{l}{\textbf{97.67\tiny±0.17}} & \multicolumn{1}{l}{\textbf{96.98\tiny±0.31}} \\
    \midrule
    \multicolumn{1}{c}{\multirow{7}[1]{*}{\parbox{1.25cm}{Fashion-\\ MNIST}}} & LLPFC & 88.40\tiny±0.23 & \multicolumn{1}{l}{85.85\tiny±0.03} & \multicolumn{1}{l}{73.63\tiny±0.48} & \multicolumn{1}{l}{28.36\tiny±0.76} & \multicolumn{1}{l}{20.03\tiny±0.47} & \multicolumn{1}{c}{\multirow{7}[1]{*}{96.39\tiny±0.02}}\\
      & DLLP & 94.20\tiny±0.02 & \multicolumn{1}{l}{93.70\tiny±0.39} & \multicolumn{1}{l}{93.18\tiny±0.22} & \multicolumn{1}{l}{91.70\tiny±0.21} & \multicolumn{1}{l}{89.62\tiny±0.46} \\
      & LLP-VAT & 94.69\tiny±0.20 & \multicolumn{1}{l}{94.17\tiny±0.16} & \multicolumn{1}{l}{93.25\tiny±0.18} & \multicolumn{1}{l}{92.30\tiny±0.13} & \multicolumn{1}{l}{89.51\tiny±0.51} \\
      & ROT & 94.25\tiny±0.17 & \multicolumn{1}{l}{93.68\tiny±0.22} & \multicolumn{1}{l}{92.53\tiny±0.46} & \multicolumn{1}{l}{91.84\tiny±0.19} & \multicolumn{1}{l}{90.14\tiny±0.31} \\
      & SoftMatch & {95.85\tiny±0.22} & \multicolumn{1}{l}{\textbf{95.86\tiny±0.25}} & \multicolumn{1}{l}{95.18\tiny±0.21} & \multicolumn{1}{l}{\textbf{94.73\tiny±0.20}} & \multicolumn{1}{l}{93.29\tiny±0.19} \\
      & FLMm* & - & \multicolumn{1}{c}{-} & \multicolumn{1}{c}{-} & \multicolumn{1}{c}{-} & \multicolumn{1}{c}{-} \\
      & \baby & \textbf{96.93\tiny±0.23} & \multicolumn{1}{c}{{95.78\tiny±0.15}} & \multicolumn{1}{c}{\textbf{95.27\tiny±0.13}} & \multicolumn{1}{c}{{94.19\tiny±0.14}} & \multicolumn{1}{c}{\textbf{93.51\tiny±0.49}} \\
        \midrule
    \multicolumn{1}{c}{\multirow{7}[1]{*}{MiniImageNet}} & LLPFC & - & \multicolumn{1}{c}{-} & \multicolumn{1}{c}{-} & \multicolumn{1}{c}{-} & \multicolumn{1}{c}{-} & \multicolumn{1}{c}{\multirow{7}[1]{*}{73.95\tiny±0.22}} \\
      & DLLP & 64.53\tiny±0.41 & \multicolumn{1}{l}{55.37\tiny±0.38} & \multicolumn{1}{l}{27.57\tiny±0.20} & \multicolumn{1}{l}{9.06\tiny±0.14} & \multicolumn{1}{l}{3.40\tiny±0.14} \\
      & LLP-VAT & 64.17\tiny±0.34 & \multicolumn{1}{l}{54.36\tiny±0.29} & \multicolumn{1}{l}{30.96\tiny±0.24} & \multicolumn{1}{l}{9.69\tiny±0.17} & \multicolumn{1}{l}{4.90\tiny±0.09} \\
      & ROT & 67.02\tiny±0.34 & \multicolumn{1}{l}{27.49\tiny±0.38} & \multicolumn{1}{l}{6.01\tiny±0.30} & \multicolumn{1}{l}{3.50\tiny±0.10} & \multicolumn{1}{l}{1.75\tiny±0.13} \\
      & SoftMatch & 2.02\tiny±0.23 & \multicolumn{1}{l}{1.86\tiny±0.24} & \multicolumn{1}{l}{1.95\tiny±0.20} & \multicolumn{1}{l}{1.72\tiny±0.33} & \multicolumn{1}{l}{1.87\tiny±0.25} \\
      & FLMm* & - & \multicolumn{1}{c}{-} & \multicolumn{1}{c}{-} & \multicolumn{1}{c}{-} & \multicolumn{1}{c}{-} \\
      & \baby & \textbf{70.26\tiny±0.26} & \multicolumn{1}{l}{\textbf{59.81\tiny±0.21}} & \multicolumn{1}{l}{\textbf{37.51\tiny±0.16}} & \multicolumn{1}{l}{\textbf{16.91\tiny±0.15}} & \multicolumn{1}{l}{\textbf{7.46\tiny±0.08}} \\
    \bottomrule
    \end{tabular}
    \vspace{-3pt}
\end{table*}

\subsection{Implementation Details}
\label{Sec:details}

For SVHN, Fashion-MNIST, and CIFAR-10, we used the WRN-28-2~\citep{zagoruyko2016wide} architecture as the encoder, and for CIFAR-100, the WRN-28-8. For MiniImageNet, we employed the ResNet-18 architecture. The classifier consists of a single linear layer. Each training step utilizes a batch size equal to the bag size multiplied by the number of bags, amounting to 1024. Models are trained using Stochastic Gradient Descent (SGD)~\citep{polyak1964some}, with a momentum of 0.9 and a weight decay of 5e-4 for WRN-28-2,  1e-3 for WRN-28-8, and 1e-4 for ResNet-18. The initial learning rate is set to \( \eta_0 = 0.03 \) for all datasets, except for MiniImageNet, where it is set to \( \eta_0 = 0.05 \), following a cosine learning rate decay schedule~\citep{loshchilov2017sgdr} as \( \eta = \eta_0 \cos\left(\frac{7\pi k}{16K}\right) \), where \( k \) is the current training step and \( K \) is the total number of steps, set to \( 2^{20} \). We train our models for \( 1024 \) epochs. Weak augmentations include random horizontal flips and random cropping. For SVHN, in accordance with~\citep{Sohn2020FixMatch}, horizontal flips are replaced with random translations up to 12.5\% in both directions. Strong augmentations utilize RandAugment~\citep{cubuk2020randaugment}. The setting for fully supervised learning is the same as mentioned above. 

\paragraph{Baselines.} 
A) LLPFC~\citep{zhang2022learning}: This method offers a theoretically grounded approach to LLP by reducing it to learning with label noise, employing the forward correction (FC) loss.
B) DLLP~\citep{ardehaly2017co}: Traditional LLP methods primarily rely on the model's output predictions to estimate the proportions of bags.
C) LLP-VAT~\citep{tsai2020learning}: This method is based on consistency regularization, where the same sample under different data augmentations should yield similar prediction results.
D) ROT~\citep{dulac-arnold2019deep}: This method uses optimal transport to generate pseudo-labels for instance-level training combined with bag-level proportion loss. To ensure fairness, we employed the same data augmentation methods used in our approach for instance-level training.
E) SoftMatch~\citep{Chen2023SoftMatch}: 
This method is one of the most important adaptive weighting techniques in weakly-supervised learning. We adopt the truncated Gaussian function, proposed in~\citep{Chen2023SoftMatch}, as the instance-level adaptive weight.  
F) FLMm~\citep{yang2021two}: This method entails pre-training with contrastive learning using a larger encoder, ResNet-50, followed by fine-tuning utilizing the FLMm approach.
$\ast$) Fully supervised: Our fully-supervised approach uses instance-level ground-truth labels for training. 

 All comparison methods except FLMm, including the fully-supervised approach, are reproduced by us. On the same dataset, we use the same experimental settings. In our method, hyper parameter settings are \(\lambda = 0.5\), \(\beta_{\mathrm{b}} = 1\), \(\beta_{\mathrm{i}} = 1\), and specially for CIFAR-100 and Mini-ImageNet, \(\beta_{\mathrm{b}} = 5\), \(\beta_{\mathrm{i}} = 5\). 

 \begin{table*}[ht]
 \caption{Classification accuracy (mean±std) of ablation experiments with respect to weight. The highest accuracy is highlighted in bold.}
  \label{tab:ablation}%
  \centering
    \begin{tabular}{c|cc|lllll}
    \toprule
    \multirow{2}[1]{*}{Dataset} & \multicolumn{2}{c|}{Weight} & \multicolumn{5}{c}{Bag Size} \\
\cmidrule{2-8}      & $\omega^{\mathrm{b}}$ & $\omega^{\mathrm{i}}$ & \multicolumn{1}{c}{16} & \multicolumn{1}{c}{32} & \multicolumn{1}{c}{64} & \multicolumn{1}{c}{128} & \multicolumn{1}{c}{256} \\
    \midrule
    \multirow{4}[1]{*}{CIFAR-10} & \xmark & \xmark & 94.60\tiny±0.23 & 94.01\tiny±0.31 & 93.61\tiny±0.30 & 86.58\tiny±0.54 & 51.90\tiny±0.35 \\
      & \cmark & \xmark & 94.63\tiny±0.33 & 94.51\tiny±0.29 & 94.21\tiny±0.34 & 92.85\tiny±0.35 & 78.03\tiny±0.25 \\
      & \xmark & \cmark & \textbf{95.00\tiny±0.08} & 94.46\tiny±0.24 & 94.31\tiny±0.18 & 92.23\tiny±0.27 & 65.54\tiny±0.24 \\
      & \cmark & \cmark & 94.96\tiny±0.13 & \textbf{95.00\tiny±0.11} & \textbf{94.58\tiny±0.21} & \textbf{93.64\tiny±0.20} & \textbf{92.88\tiny±0.53} \\
    \midrule
    \multirow{4}[1]{*}{CIFAR-100} & \xmark & \xmark & 77.54\tiny±0.37 & 8.91\tiny±2.49 & 3.05\tiny±0.94 & 2.23\tiny±0.87 & 1.96\tiny±0.95 \\
      & \cmark & \xmark & 76.68\tiny±0.46 & 75.40\tiny±0.52 & 72.97\tiny±0.23 & 65.09\tiny±0.13 & 21.59\tiny±0.09 \\
      & \xmark & \cmark & 77.97\tiny±0.86 & 76.86\tiny±0.32 & 70.99\tiny±0.74 & 69.48\tiny±0.19 & 2.89\tiny±1.09 \\
      & \cmark & \cmark & \textbf{78.65\tiny±0.28} & \textbf{77.30\tiny±0.50} & \textbf{76.52\tiny±0.23} & \textbf{72.21\tiny±0.37} & \textbf{23.56\tiny±2.13} \\
    \bottomrule
    \end{tabular}
\end{table*}

\subsection{Results and Analysis}

\paragraph{\baby achieves SOTA results.}
We present results on five benchmark datasets in Table~\ref{tab:result}. We observe that \baby significantly outperforms the majority of baseline methods across all benchmark datasets and for every bag size. Compared with FLMm, which utilizes a stronger pre-trained encoder, our method's performance is only lower on CIFAR-100 with a bag size of 256. In other bag sizes, our method significantly surpasses FLMm, particularly on CIFAR-100 with bag sizes of 16 and 32, where we exceed their performance by 12.49\% and 11.71\%, respectively. For methods other than FLMm, our results exhibit a more gradual decline with increasing bag sizes. For instance, as the bag size increases from 16 to 256 on CIFAR-10, \baby's performance drops by only 2.08\%, whereas other methods show a decline of over 40\%. 
Compared to ROT, which also uses pseudo-label self-training, we consistently achieve significantly better results in all cases. 
SoftMatch performs well in certain cases, particularly when the bag size is small, such as 16 on CIFAR-10/100. However, as the bag size increases, its performance drops significantly. For example, when the bag size is 128 on CIFAR-100, the accuracy falls to only 2.12\%. Although SoftMatch uses a similar technique to our method, the adaptive weighting it employs fails to address the issue of over-smoothing.
More importantly, \baby achieves results comparable to those of fully supervised learning models. For example, at a bag size of 16, \baby's performance on simpler datasets like Fashion-MNIST and SVHN matches that of fully supervised learning. On more complex datasets such as CIFAR-10 and CIFAR-100, there's only a slight performance degradation, lower than 2\%.
For the more challenging MiniImageNet dataset, \baby achieves the highest accuracy across all bag sizes.

\begin{figure}[tbp]
    \centering
    \subcaptionbox{CIFAR-10}{\includegraphics[scale=0.15]{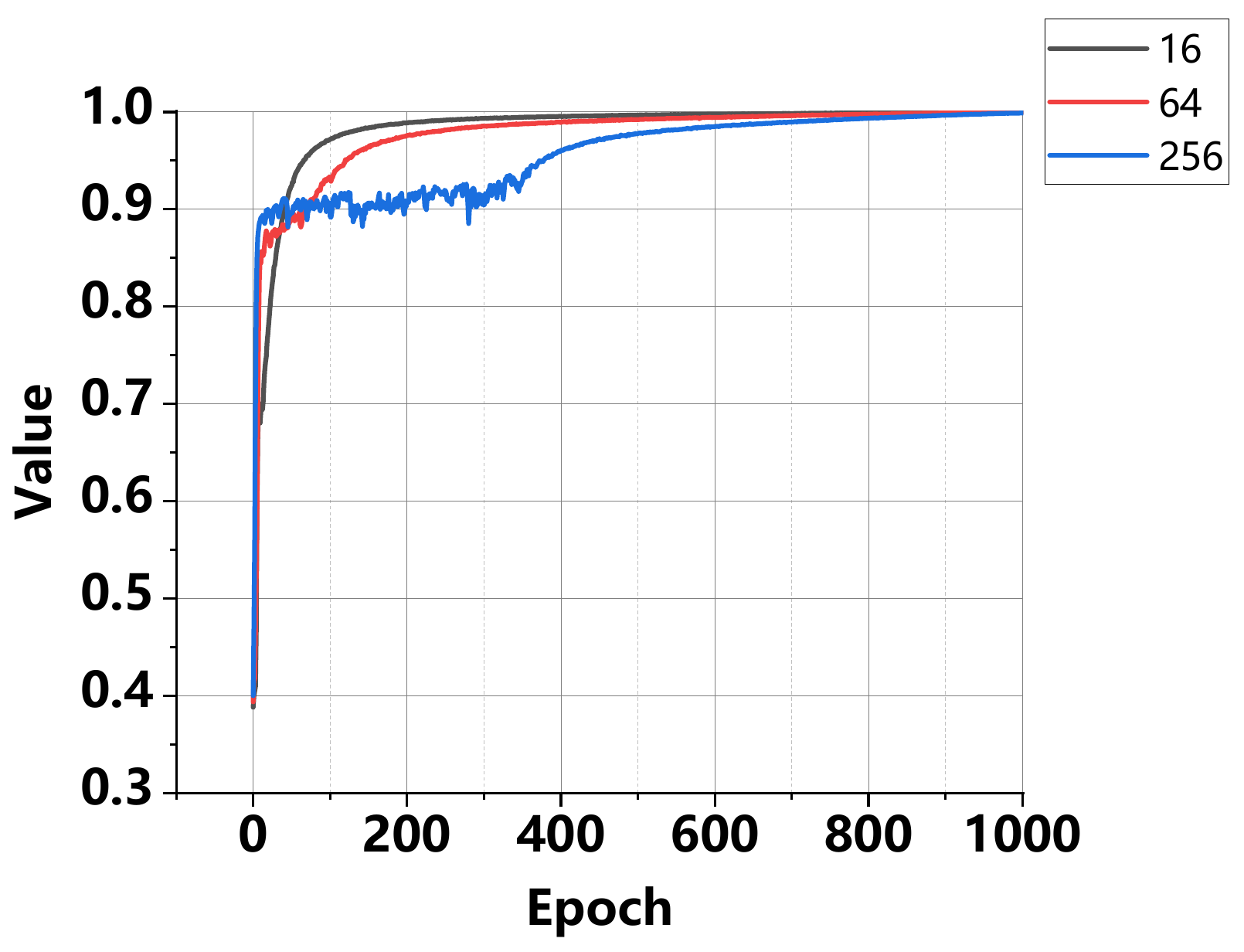}}
    \subcaptionbox{CIFAR-100}{\includegraphics[scale=0.15]{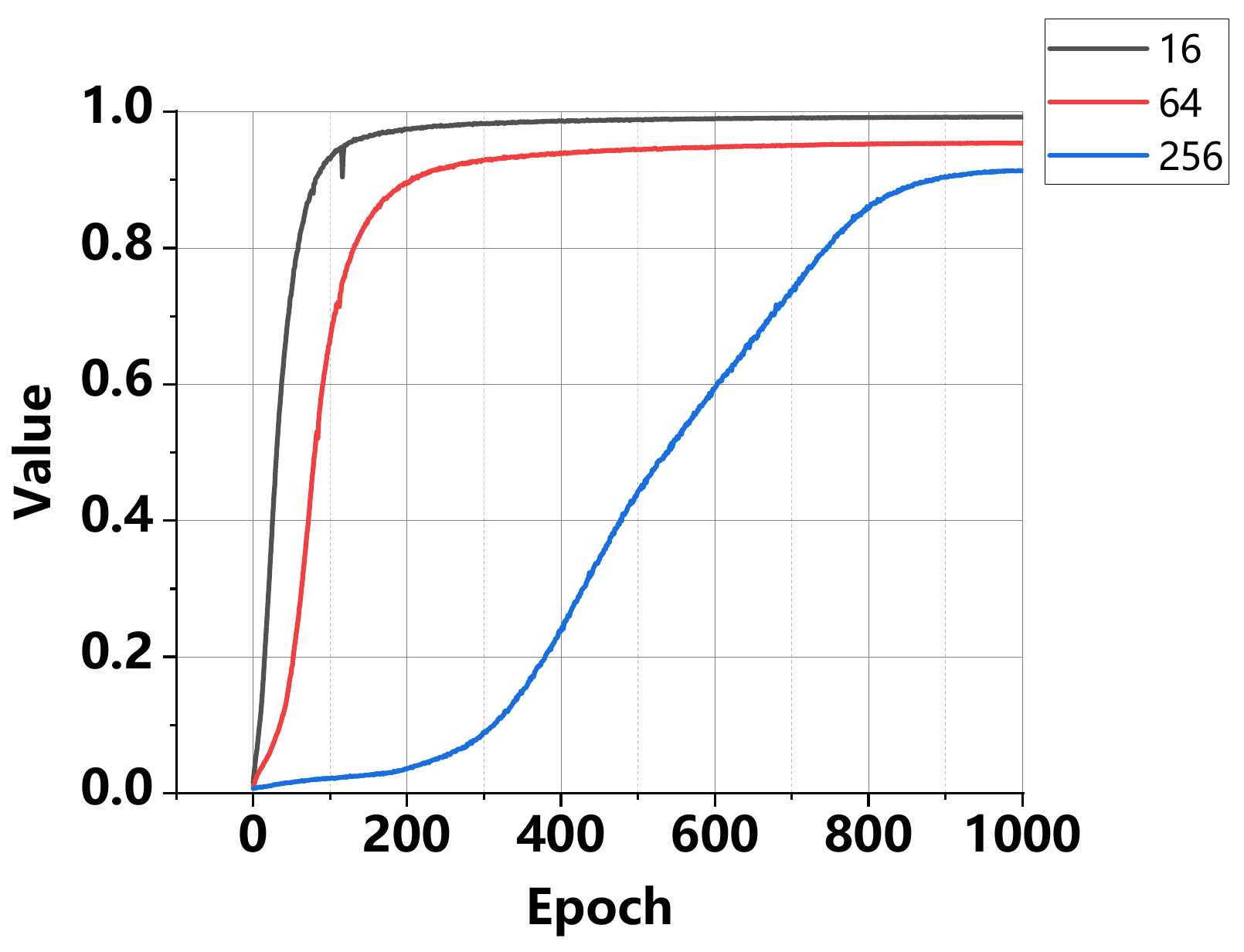}}
    \caption{Evolution of adaptive weights across training epochs for bag sizes 16, 64 and 256 on CIFAR-10 and CIFAR-100.}
    \label{fig:epoch}
\end{figure}

\paragraph{Variations in adaptive weighting during training. }
\label{variations in adaptive weighting}
As shown in Fig.~\ref{fig:epoch}, the adaptive weights in \baby increase progressively throughout the training process, potentially reaching high values to fully leverage the information from pseudo-labels. Simultaneously, \baby effectively measures the confidences of pseudo-labels, efficiently minimizing the interference of meaningless pseudo-labels in the training process and achieving state-of-the-art results.

\begin{figure*}[htbp]
    \centering
    \subcaptionbox{DLLP features}{\includegraphics[scale=0.073]{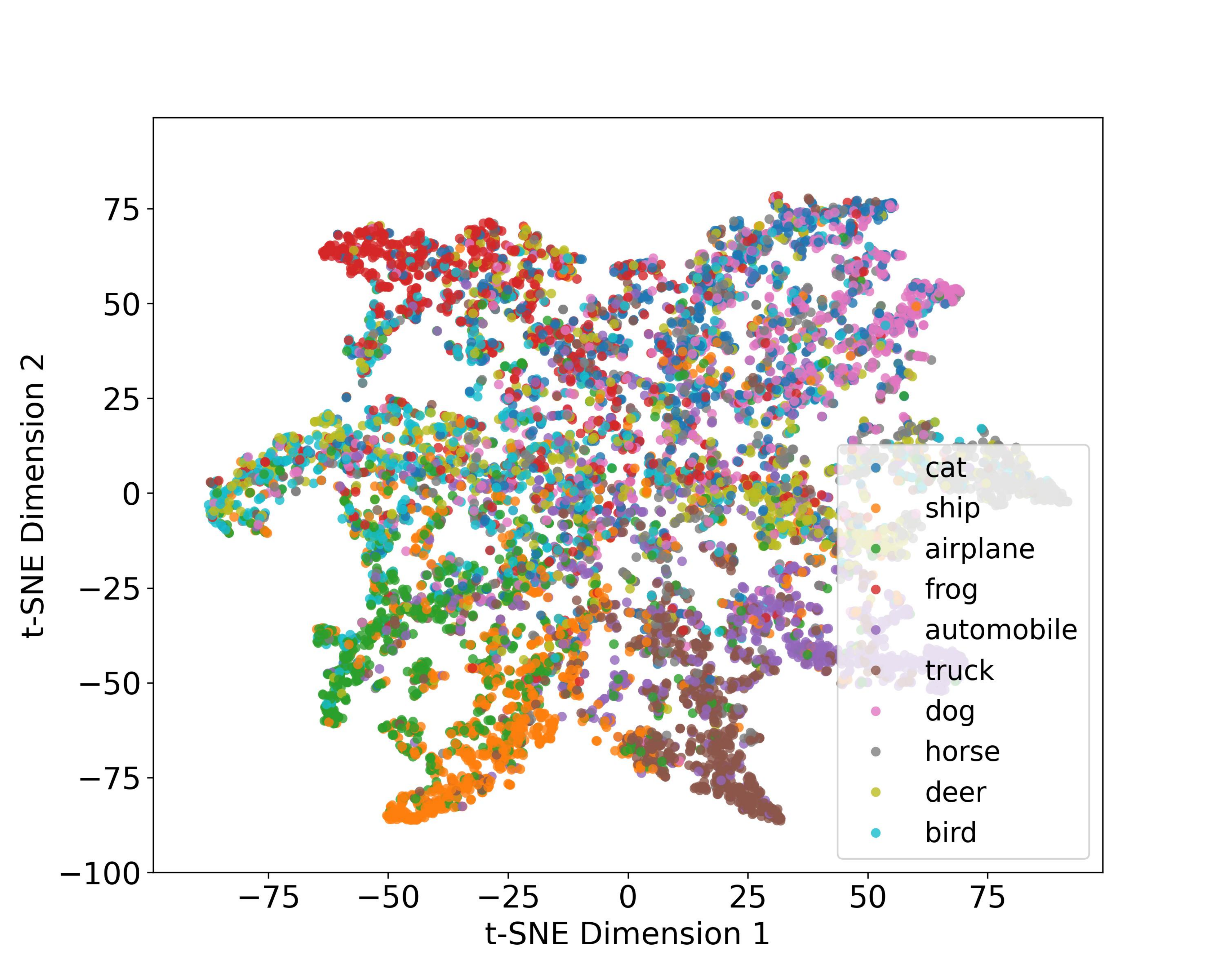}\label{fig:t-sne-DLLP}}
    \subcaptionbox{ROT features}{\includegraphics[scale=0.073]{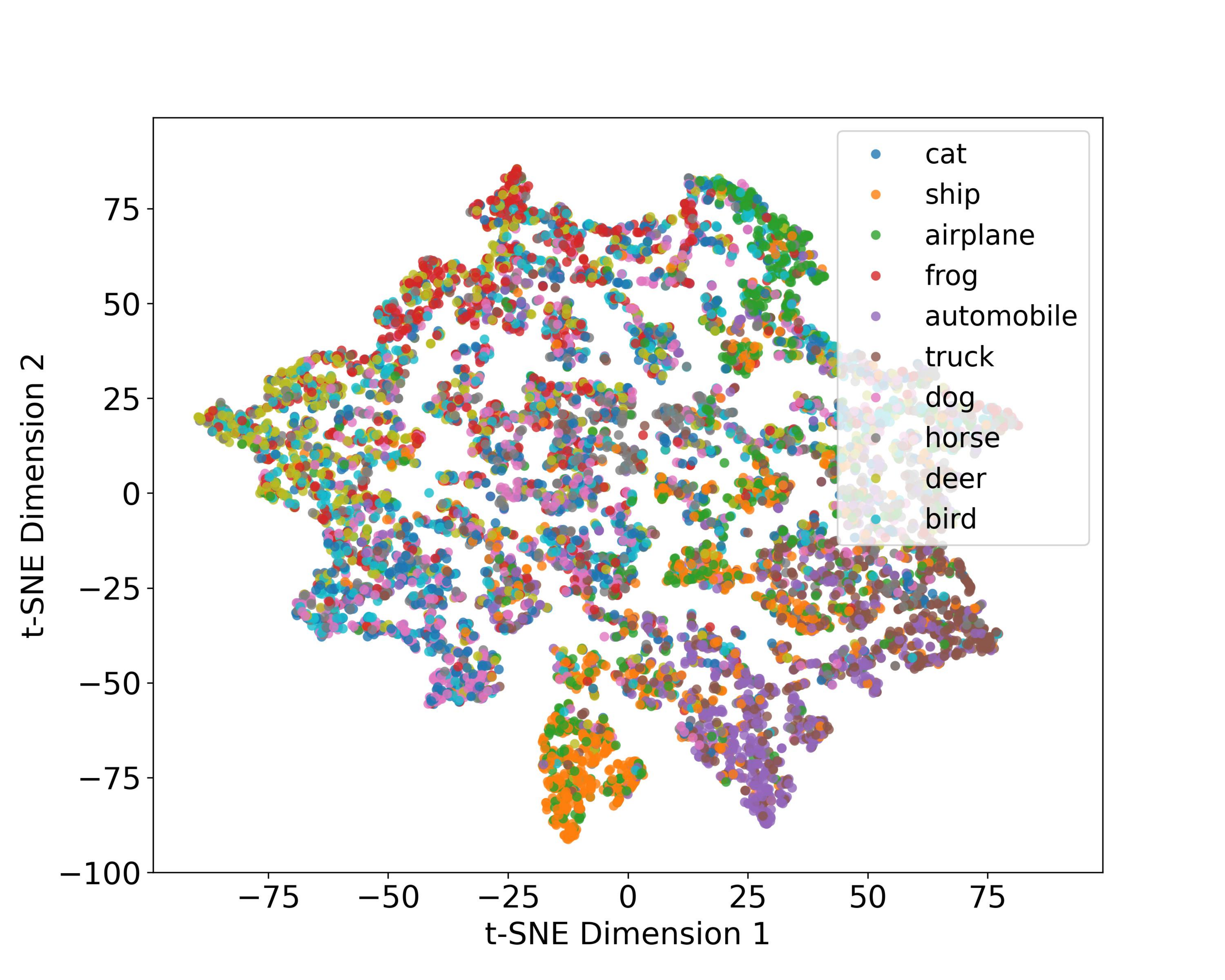}\label{fig:t-sne-ROT}}
    \subcaptionbox{\baby features(ours)}{\includegraphics[scale=0.073]{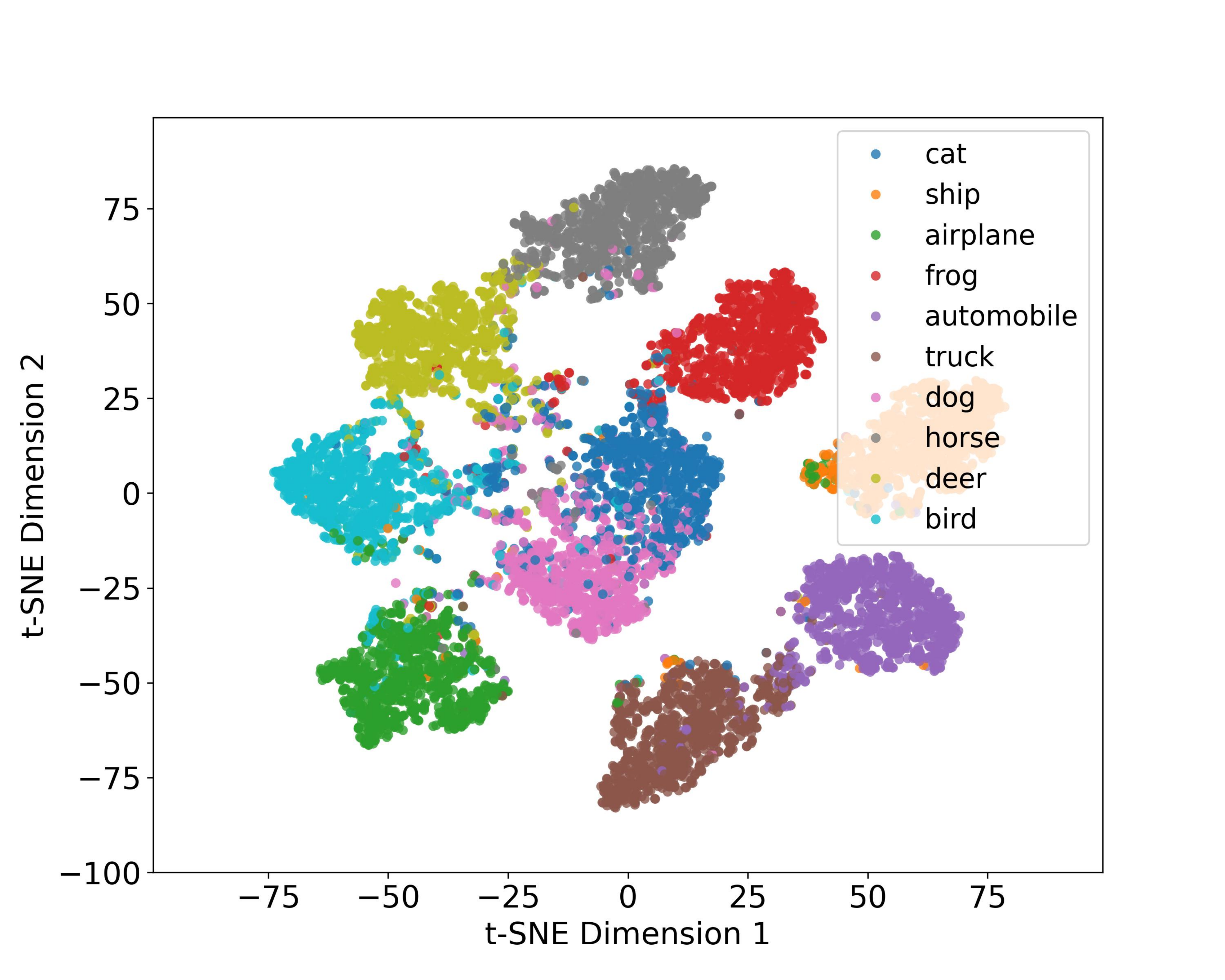}\label{t-sne-LLPAHIL}}
    \caption{t-SNE visualization of learned feature spaces on CIFAR-10 with bag size 256. Distinct colors correspond to different classes, where (a) DLLP and (b) ROT baselines exhibit overlapping clusters, while (c) our method \baby demonstrates clearer inter-class separation and tighter intra-class clustering, indicating more discriminative feature learning.}
    \label{fig:t-sne}
\end{figure*}

\paragraph{\baby learns more distinguishable representations.}
We visualize the image representations generated by the feature encoder using t-SNE~\citep{van2008visualizing} in Fig.~\ref{fig:t-sne}. Different colors correspond to different ground-truth class labels. We use the CIFAR-10 dataset with a bag size 256 and compare the t-SNE embeddings of three classic methods. For the traditional method DLLP, training the model solely with the bag-level proportion loss does not lead to effective representation learning. Although ROT introduces an instance-level loss through pseudo labels, it still falls short of achieving sufficient representation learning. In contrast, \baby generates well-separated clusters and more distinct representations, demonstrating its superior ability to learn high-quality representations.

\subsection{Ablation Study}

In this section, we perform an ablation study to assess the influence of two central elements of \baby: bag-level adaptive weight $\omega^{\rm b}$ and instance-level adaptive weight $\omega^{\rm i}$. Experiments were conducted on the CIFAR-10 and CIFAR-100 datasets, with results presented in Table~\ref{tab:ablation}. 
The first row shows that removing both adaptive weights. With smaller bag sizes and fewer classes, the pseudo-labels generated by the model were relatively accurate, and the classification performance did not decline significantly, as seen with bag sizes from 16 to 64 on CIFAR-10. However, for larger bag sizes, such as 256, accuracy plummeted to merely 51.90\%. On the more complex CIFAR-100, with a greater number of classes, the model's performance deteriorated substantially. On CIFAR-100, we observe that the model's performance drops sharply between adjacent bag sizes. The large number of classes and the reduction in supervisory information make the model more prone to generating inaccurate pseudo-labels, a phenomenon also observed in other weakly supervised domains. Without adaptive weights, the model is heavily impacted by incorrect pseudo-labels, leading to a significant performance drop.
The second and third rows indicate that the model's accuracy decreases when the instance-level and bag-level adaptive weights are removed, respectively. This degradation intensifies with increasing bag sizes, highlighting that reliance on a single weight (bag-level or instance-level) inadequately captures pseudo-label~confidence.

\subsection{Parameter Sensitivity}

\begin{figure}[tbp]
    \centering
    \newcommand{\uniformsize}{\scriptsize} 
    \begin{minipage}{\linewidth}
        \centering
        \subcaptionbox{CIFAR-10}{\includegraphics[scale=0.205]{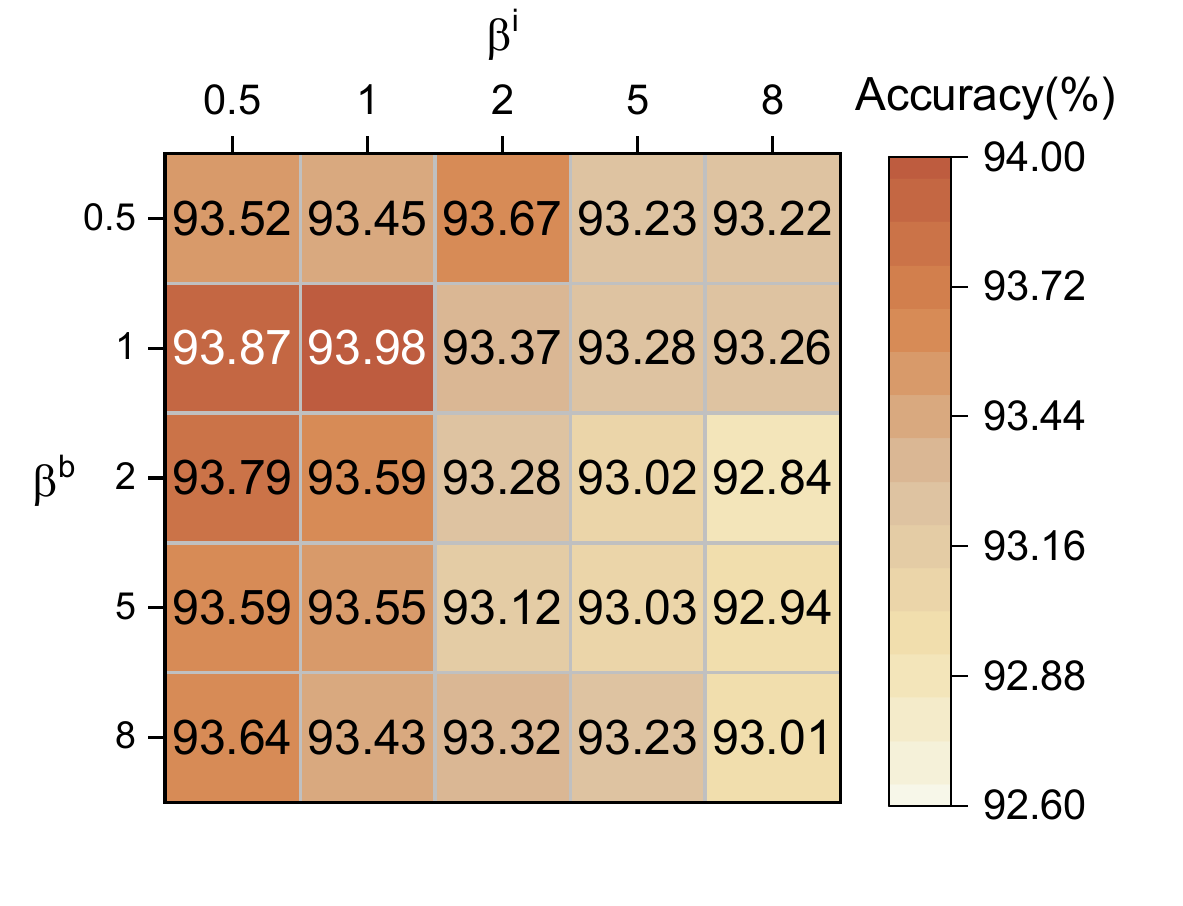}}
        \subcaptionbox{CIFAR-100}{\includegraphics[scale=0.205]{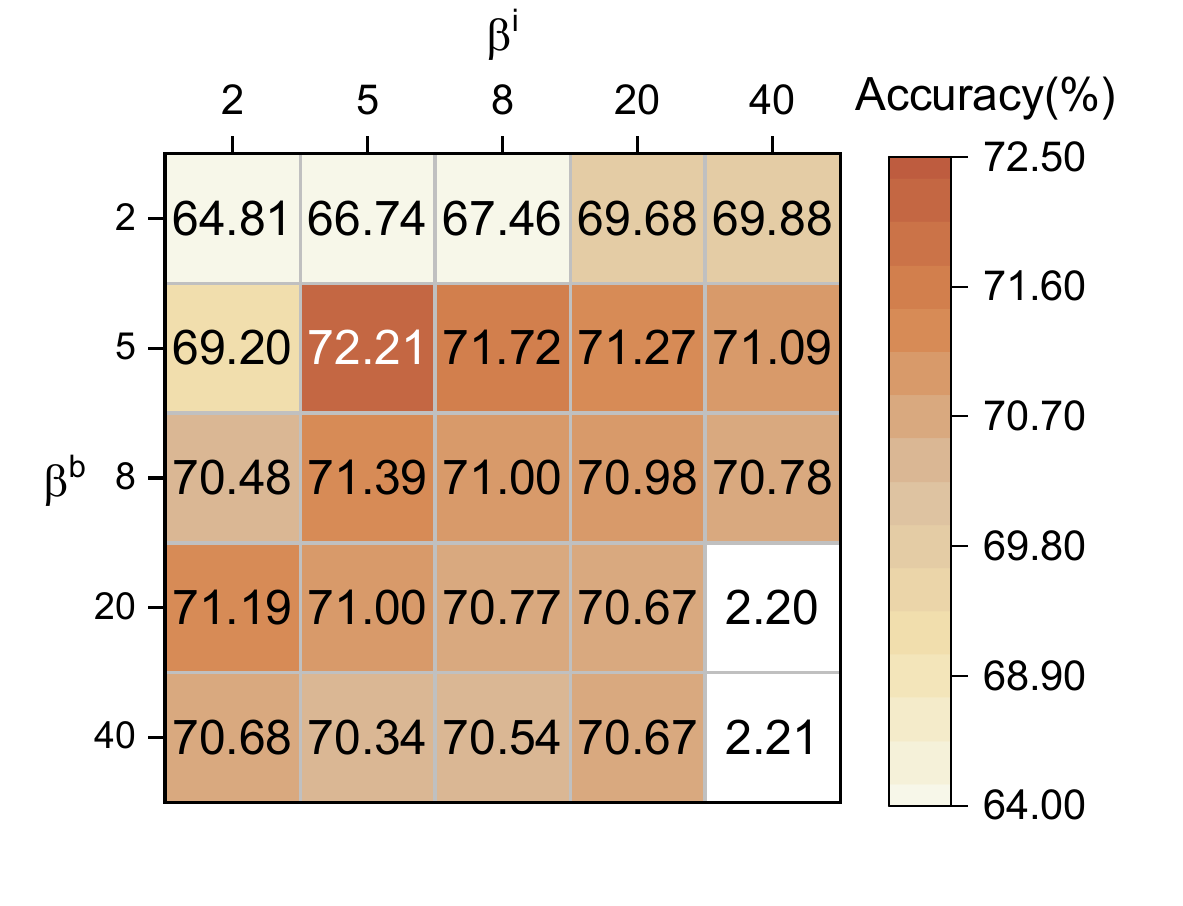}}
        \caption{The plots of joint sensitivity tests for hyper parameters $\beta_{\rm{b}}$ and $\beta_{\rm{i}}$ on CIFAR-10 and CIFAR-100 with a bag size 128. }    
        \label{fig:sensitivity}
    \end{minipage}
\end{figure}

In this section, we investigate how \(\beta_{\mathrm{b}}\) and \(\beta_{\mathrm{i}}\), two key hyper parameters in \baby that control the mapping functions at bag-level and instance-level respectively, influence model performance. We conduct experiments on CIFAR-10 and CIFAR-100 with a bag size 128 (Fig.~\ref{fig:sensitivity}). For the 100-class CIFAR-100 task, the best result is achieved with \(\beta_{\mathrm{b}}=5\) and \(\beta_{\mathrm{i}}=5\). For the 10-class task, represented by CIFAR-10, the optimal result is obtained with \(\beta_{\mathrm{b}}=1\) and \(\beta_{\mathrm{i}}=1\). We selected our hyper parameters \(\beta_{\mathrm{b}}\) and \(\beta_{\mathrm{i}}\) based on these tests. When \(\beta_{\mathrm{b}}\) and \(\beta_{\mathrm{i}}\) are small, the weights are highly sensitive to changes in entropy, and some accurate pseudo-labels may be assigned low weights, preventing the model from fully utilizing them. For example, when \(\beta_{\mathrm{b}}\) = 2 and \(\beta_{\mathrm{i}}\) = 2, the accuracy on CIFAR-100 is only 64.81\%. Conversely, when \(\beta_{\mathrm{b}}\) and \(\beta_{\mathrm{i}}\) are too large, many smooth but inaccurate pseudo-labels receive high weights, which hampers classifier training. For instance, with \(\beta_{\mathrm{b}}\) = 40 and \(\beta_{\mathrm{i}}\) = 40, the model fails to converge on CIFAR-100.

\section{Conclusion}
\label{conclusion}

In this paper, we evaluated the problem of imprecise pseudo-labels in existing LLP methods. Addressing these challenges, we introduced \baby, a novel approach that measures pseudo-label confidence using both bag-level and instance-level entropy values, and then adaptively adjusts the pseudo-label weight for each instance. Our extensive experiments showed that \baby achieves state-of-the-art performance and remains effective even with larger bag sizes. We also conducted comprehensive ablation studies, confirming the effectiveness of each weight component in our method. 
\section{Limitations}
\label{Sec:limiations}

The major limitation of \baby is that the hyper parameters of the adaptive weight is relatively sensitive to the datasets with various class numbers. Besides, our test dataset is considered classical and easily accessible, yet it has not been evaluated on larger, more complex datasets. 
{
    \small
    \bibliographystyle{ieeenat_fullname}
    \bibliography{main}
}
\end{document}